\documentclass[letterpaper]{article} 
\usepackage{aaai25}  
\usepackage{times}  
\usepackage{helvet}  
\usepackage{courier}  
\usepackage[hyphens]{url}  
\usepackage{graphicx} 
\usepackage{booktabs}       
\usepackage{amsfonts}       
\usepackage{natbib}  
\usepackage{caption} 
\usepackage{algorithm}
\usepackage{algorithmic}
\usepackage{amsmath}
\usepackage[table,xcdraw]{xcolor}
\usepackage{makecell}
\usepackage{multirow}
\usepackage{multicol}
\usepackage{amsmath}
\usepackage{amssymb}
\usepackage{adjustbox}
\usepackage{xcolor}
\usepackage[hang,flushmargin]{footmisc}

\usepackage{newfloat}
\usepackage{listings}
\DeclareCaptionStyle{ruled}{labelfont=normalfont,labelsep=colon,strut=off} 
\floatstyle{ruled}
\newfloat{listing}{tb}{lst}{}
\floatname{listing}{Listing}
\def\correspondingauthor{%
	\ifnum\value{correspondingfn}=0%
	\footnote{Corresponding authors: Chen Gong, Shuo Chen.}%
	\setcounter{correspondingfn}{\value{footnote}}%
	\else%
	\footnotemark[\value{correspondingfn}]%
	\fi%
}

\newcounter{correspondingfn}

\title{Hybrid Data-Free Knowledge Distillation}
\author{
    Jialiang Tang$^{1, 2, 3}$, Shuo Chen$^{4}$\correspondingauthor{}, Chen Gong$^{5}$\correspondingauthor{}
}
\affiliations{
    \textsuperscript{\rm 1}School of Computer Science and Engineering, Nanjing University of Science and Technology, China\\
    \textsuperscript{\rm 2}Key Laboratory of Intelligent Perception and Systems for High-Dimensional Information of Ministry of Education, China\\
    \textsuperscript{\rm 3}Jiangsu Key Laboratory of Image and Video Understanding for Social Security, China \\
    \textsuperscript{\rm 4}Center for Advanced Intelligence Project, RIKEN, Japan \\
    \textsuperscript{\rm 5}Department of Automation, Institute of Image Processing and Pattern Recognition, Shanghai Jiao Tong University, China \\
    tangjialiang@njust.edu.cn, shuo.chen.ya@riken.jp, chen.gong@sjtu.edu.cn
}

\usepackage{bibentry}

\begin{document}

\maketitle

\begin{abstract}
	Data-free knowledge distillation aims to learn a compact student network from a pre-trained large teacher network without using the original training data of the teacher network. Existing collection-based and generation-based methods train student networks by collecting massive real examples and generating synthetic examples, respectively. However, they inevitably become weak in practical scenarios due to the difficulties in gathering or emulating sufficient real-world data. To solve this problem, we propose a novel method called \textbf{H}ybr\textbf{i}d \textbf{D}ata-\textbf{F}ree \textbf{D}istillation (HiDFD), which leverages only a small amount of collected data as well as generates sufficient examples for training student networks. Our HiDFD comprises two primary modules, \textit{i.e.}, the teacher-guided generation and student distillation. The teacher-guided generation module guides a Generative Adversarial Network (GAN) by the teacher network to produce high-quality synthetic examples from very few real-world collected examples. Specifically, we design a feature integration mechanism to prevent the GAN from overfitting and facilitate the reliable representation learning from the teacher network. Meanwhile, we drive a category frequency smoothing technique via the teacher network to balance the generative training of each category. In the student distillation module, we explore a data inflation strategy to properly utilize a blend of real and synthetic data to train the student network via a classifier-sharing-based feature alignment technique. Intensive experiments across multiple benchmarks demonstrate that our HiDFD can achieve state-of-the-art performance using 120 times less collected data than existing methods. Code is available at https://github.com/tangjialiang97/HiDFD.
\end{abstract}
\section{Introduction}
\label{sec_intro}
The success of Deep Neural Networks (DNNs)~\cite{he2016deep,haodata} is usually accompanied by significant computational and storage demands, which hinders their deployment on practical resource-limited devices. Knowledge Distillation (KD)~\cite{Hinton2015Distilling,miles2024understanding} has served as an effective compression technology that transfers knowledge from a complex pre-trained teacher network to improve the performance of a lightweight student network. However, in practice, the training data of the teacher network is usually inaccessible due to privacy concerns and only the pre-trained teacher network itself can be used to learn the student network. This is because users may prefer sharing a pre-trained black box DNN rather than disclosing their sensitive data. In such cases, vanilla KD methods can hardly train a reliable student network owing to the absence of original training data. To address this issue, various Data-Free Knowledge Distillation (DFKD) approaches~\cite{binici2022robust,chen2019data,chen2021learning,tang2023distribution} have been developed to enable training the student network without using any original data. 

Among existing DFKD methods, collection-based approaches~\cite{chen2021learning,tang2023distribution} can achieve satisfactory performance by amassing numerous real examples to train the student network. However, it is still difficult for the collection-based methods to train a reliable student network in practical tasks, \textit{e.g.}, medical image classification because gathering sufficient training examples can be challenging. On the other hand, generation-based methods~\cite{yin2020dreaming,chen2019data} leverage the teacher network to guide a generative model~\cite{creswell2018generative} in producing fake examples, thereby successfully training the student network without reliance on real examples. Nevertheless, the synthesized examples may exhibit low quality in the absence of real data supervision, leading to suboptimal student performance, especially for many challenging recognition tasks on ImageNet~\cite{deng2009imagenet}. The inherent constraints of both collection-based and generation-based DFKD methods prompt an essential question: \textit{Can we train an effective generative model only using a small number of collected examples and then learn reliable student networks with the hybrid data comprising both collected and synthetic examples?}

To answer the above question under the practical data-free distillation scenario, we need a generative model that not only possesses powerful generative capabilities but also has the ability to acquire valuable knowledge from the teacher network. Recent studies~\cite{cui2023kd, rangwani2021class} suggest that the Generative Adversarial Network (GAN)~\cite{mirza2014conditional} can easily learn from pre-trained models and then generate high-quality synthetic examples, so we employ this great approach as our generative module. The standard GAN consists of a generator and a discriminator trained in an adversarial manner, where the generator attempts to produce fake examples to deceive the discriminator while the discriminator strives to distinguish between real and fake examples. However, the collected data in practice tasks like medical image classification has two inherent characteristics that may impede the training of the GAN, namely: 1) \textit{Limited data quantity}, as capturing medical images requires expensive and complex equipment; and 2) \textit{Imbalanced class distribution}, where certain diseases (\textit{e.g.}, ``vascular lesions'') are more rarely than others (\textit{e.g.}, ``nevus''). When training on the collected data with limited examples and imbalanced class distribution, the discriminator is susceptible to overfitting~\cite{huang2022masked,jiang2021deceive}. It implies that the discriminator tends to memorize all real examples and almost perfectly distinguish them from fake examples, resulting in the disappearance of the gradient for the generator. Moreover, the generator training is dominated by a few classes occupying the majority of examples, which prevents it from generating diverse examples. Therefore, it is critical to overcome the overfitting issue of discriminator and data imbalance issue of generator when training with scarce collected examples.

In this paper, we propose a novel approach called \textbf{H}ybr\textbf{i}d \textbf{D}ata-\textbf{F}ree \textbf{D}istillation (HiDFD), which learns reliable student networks on the hybrid data comprising synthetic examples and very few real collected examples. Our HiDFD is composed of two pivotal modules of teacher-guided generation and student distillation. In the teacher-guided generation module, we aim to solve the critical issues in the GAN mentioned above, and thus generating high-quality synthetic examples. Specifically, we propose a feature integration mechanism to aggregate the features of both the collected and synthetic examples between the teacher network and GAN. Such an integration mechanism not only mitigates the overfitting of the discriminator, which forcibly distinguishes those closely resembling examples, but also transfers valuable representations to guide the discriminator to capture category dependencies. Meanwhile, we also develop a new technique called category frequency smoothing to alleviate the imbalanced training of the generator. In the student distillation module, we develop a data inflation operation to adjust the contribution of collected examples among the hybrid data when training the student network. Finally, we design a classifier-sharing-based strategy to closely align the features of student network with those of teacher network to enhance student performance. Thanks to effectively transferring knowledge from the teacher network to both the GAN and student network, our HiDFD can successfully train reliable student networks using very few collected real-world examples. The contributions of our HiDFD are summarized as follows:
\begin{itemize}
	\itemsep=1pt
	\item By considering the difficulties in gathering or emulating real-world data, we propose a novel data-free distillation method called HiDFD, which only requires a small number of collected data to generate high-quality synthetic examples for training the student network.
	\item We design a teacher-guided generation module to effectively tackle the critical issues of discriminator overfitting and imbalanced learning in generating synthetic examples, which empowers the distillation module to learn reliable student networks from the teacher network.
	\item Our HiDFD can achieve State-Of-The-Art (SOTA) performance using only 1/120 (5,000/600,000) of examples required by existing collection-based DFKD methods.
\end{itemize}
\section{Related Works}
In this section, we review the relevant works, including knowledge distillation and generative models.
\subsection{Knowledge Distillation}
Traditional KD methods~\cite{chen2021cross,li2023curriculum} learn a compact and reliable student network by encouraging it to mimic a variety of knowledge, \textit{i.e.}, softened logits~\cite{zhao2022decoupled}, intermediate features~\cite{chen2022knowledge}, and representation relationships~\cite{peng2019correlation}, from a large teacher network using ample original training data. However, in practical applications, these approaches might be ineffective because the original data is usually unavailable due to privacy concerns.

To address the above issue, generation-based~\cite{tran2024nayer,wang2024unpacking,wang2024confounded} and collection-based~\cite{chen2021learning,tang2023distribution} DFKD methods have been proposed to train student networks using synthetic and collected data, respectively. The generation-based methods utilize the teacher network to guide a generator in producing examples from statistics in the teacher network or random noise. However, the resulting student network still achieves suboptimal performance due to the flawed synthetic examples. Conversely, collection-based methods assume that there are numerous easily accessible examples in the real-world, and they acquire an oversized collected data (\textit{e.g.}, 600,000 examples on CIFAR10) to train the student network. In practical tasks, it is hard to gather so many examples, and thus they still fail to train reliable student networks.

In this paper, our HiDFD only utilizes a small collected data that contains fewer examples than the original data, which initially guides the GAN in training on such collected data by the teacher and then trains the student on adequate data composed of the synthetic and collected examples.
\subsection{Generative Models}
\label{sec_related_gm}
Recent advances in generative models, including Variational AutoEncoders~\cite{kingma2013auto,zhao2019infovae}, diffusion models~\cite{ho2020denoising,mei2023vidm}, and GAN~\cite{hou2021slimmable, mirza2014conditional} have significantly propelled the data generation. This paper focuses on the powerful GAN due to its ability to learn from pre-trained models~\cite{cui2023kd, rangwani2021class}. The traditional GAN~\cite{goodfellow2020generative} consists of a generator and a discriminator, where the generator produces fake examples to deceive the discriminator, and the discriminator tries to accurately distinguish between real and fake examples. Recently, Auxiliary Discriminative Classifier GAN (ADCGAN)~\cite{hou2022conditional} captures dependencies between generated examples and class labels by encouraging the discriminator to classify synthetic examples into specific categories, which effectively improves the quality of synthetic data.

In our method, we hope the GAN can produce high-quality synthetic examples that are easily classifiable, and thus training a precise student network. Therefore, we adopt ADCGAN as the foundational generative model. The ADCGAN composed of a generator $\mathcal{N}_\mathrm{G}:\mathcal{Z}\times\mathcal{Y}\rightarrow\mathcal{X}$ maps a noise-label pair $(\boldsymbol{z}, y)$ to a fake example $\mathcal{N}_\mathrm{G}(\boldsymbol{z}, y)\in \mathcal{X}$ that can be precisely predicted as $y\in\mathcal{Y}$; and a discriminator $\mathcal{N}_{\mathrm{D}}:\mathcal{X}\rightarrow\{0,1\}
$ determines whether the input example is real (\textit{i.e.}, 1) or fake (\textit{i.e.}, 0), which also has a classifier $\Psi_{\mathrm{D}}:\mathcal{X}\rightarrow\mathcal{Y}^{+}\cup\mathcal{Y}^{-}$ ($y^{+}\in \mathcal{Y}^{+}$ and $y^{-}\in\mathcal{Y}^{-}$ denote the labels for real and fake examples, respectively). Mathematically, the objective functions for the discriminator and generator in the ADCGAN are defined as $\mathcal{L}_{\mathrm{adc\_d}}$ and $\mathcal{L}_{\mathrm{adc\_g}}$, respectively, as follows:
\begin{equation}
\left\{\begin{array}{l}
\begin{aligned}
\mathcal{L}_{\mathrm{adc\_d}}=&-\mathcal{L}_{\mathrm{_d}}+\mathbb{E}_{\boldsymbol{x}, y \sim P_{X, Y}}[\log \Psi_{\mathrm{D}}(y^{+} \mid \boldsymbol{x})]
\\&+\mathbb{E}_{\boldsymbol{x},y\sim Q_{X,Y}}[\log \Psi_{\mathrm{D}}(y^{-} \mid \boldsymbol{x})],\\
\mathcal{L}_{\mathrm{adc\_g}}=&\ \ \ \mathcal{L}_{\mathrm{_g}}-\mathbb{E}_{\boldsymbol{x}, y \sim Q_{X, Y}}[\log \Psi_{\mathrm{D}}(y^{+}\mid\!\boldsymbol{x})]\\&+\mathbb{E}_{\boldsymbol{x}, y \sim Q_{X, Y}}[\log \Psi_{\mathrm{D}}(y^{-}\mid\boldsymbol{x})],
\end{aligned}
\end{array}\right.
\label{eq_gan}
\end{equation}
where $\mathcal{L}_{\mathrm{d}}=\mathbb{E}_{\boldsymbol{x} \sim P_X}[\log \mathcal{N}_{D}(\boldsymbol{x})]+\mathbb{E}_{\boldsymbol{x} \sim Q_{X}}[\log (1-\mathcal{N}_{D}(\boldsymbol{x}))]$ and $\mathcal{L}_{\mathrm{g}}=\mathbb{E}_{\boldsymbol{x} \sim Q_{X}}[\log (1-\mathcal{N}_{D}(\boldsymbol{x}))]$ are the loss functions for the standard GAN, $P$ and $Q$ denote the distribution of real collected data and fake synthetic data, respectively. $\Psi_{\mathrm{D}}(y^{+}\mid\cdot)$ (resp. $\Psi_{\mathrm{D}}(y^{-}\mid\cdot)$) denotes the probability that the input example is classified as the label $y$ and real (resp. fake) simultaneously by the following classifier of the discriminator. Formally, $\Psi_{\mathrm{D}}\left(y^{+} \mid \boldsymbol{x}\right)=\frac{\exp \left(\Psi_{\mathrm{D}}^{+}(y) \cdot \Phi_{\mathrm{D}}(\boldsymbol{x})\right)}{\sum_{\bar{y}\in\mathcal{Y}^{+}} \exp \left(\Psi_{\mathrm{D}}^{+}(\bar{y}) \cdot \Phi_{\mathrm{D}}(\boldsymbol{x})\right)+\sum_{\bar{y}\in\mathcal{Y}^{-}} \exp \left(\Psi^{-}_{\mathrm{D}}(\bar{y}) \cdot \Phi_{\mathrm{D}}(\boldsymbol{x})\right)}$, where $\Phi_{\mathrm{D}}$ represents the shared feature extractor between the original discriminator $\mathcal{N}_\mathrm{D}$ and the classifier $\Psi_{\mathrm{D}}$. $\Psi^{+}_{\mathrm{D}}$ (resp. $\Psi^{-}_{\mathrm{D}}$) captures the dependencies between the category labels and real (resp. fake) data. Notably, DeGAN~\cite{addepalli2020degan} also trains a GAN using collected data, but it still requires a large number of collected examples and can only utilize synthetic examples to train the target model.
\begin{figure*}[t]
	\centering
	\includegraphics[scale=0.295]{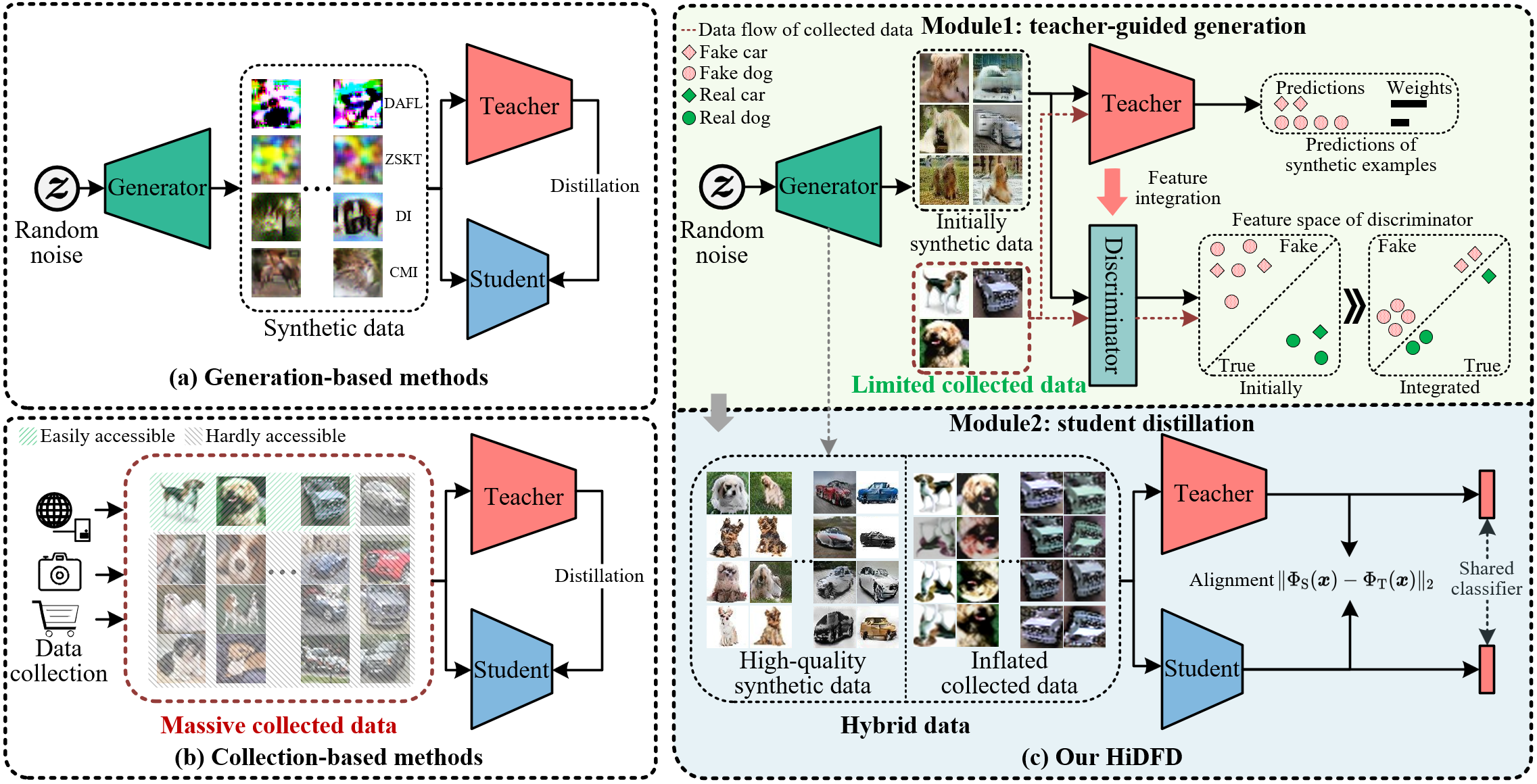}
	\caption{The diagram of (a) generation-based methods~\cite{fang2021contrastive,yin2020dreaming,chen2019data,micaelli2019zero}, (b) collection-based methods~\cite{chen2021learning,tang2023distribution}, and (c) our HiDFD. In HiDFD, the teacher-guided generation module employs the teacher network to guide the training of the GAN on limited collected data. Subsequently, the student distillation module closely aligns the features of the student network with those of the teacher network on the hybrid data comprising high-quality synthetic examples and properly inflated collected examples.}
	\label{fig_gandc}
		\vspace{-10pt}
\end{figure*}
\section{Approach}
\label{sec_app}
Data-free distillation aims to train a compact student network $\mathcal{N}_{\mathrm{S}}$ using a pre-trained teacher network $\mathcal{N}_{\mathrm{T}}$ without accessing the teacher's original training data $\mathcal{D}_{\mathrm{o}}$. Both $\mathcal{N}_{\mathrm{T}}$ and $\mathcal{N}_{\mathrm{S}}$ consist of a feature extractor $\Phi$ and classifier $\Psi$, where the subscripts $\mathrm{T}$ and $\mathrm{S}$ indicate ``teacher'' and ``student'', respectively. Existing collection-based DFKD methods~\cite{tang2023distribution} usually rely on the collected data $\mathcal{D}_{\mathrm{c}}$ with overwhelming examples searched based on the categories of the original data. Here, the data amount $|\mathcal{D}_{\mathrm{c}}|\gg|\mathcal{D}_{\mathrm{o}}|$, which is hard to satisfy in practice tasks. To overcome this limitation, we propose a more practical method that only requires a small number of collected examples for DFKD, \textit{i.e.}, the data amount $|\mathcal{D}_{\mathrm{c}}|\leq|\mathcal{D}_{\mathrm{o}}|$. To this end, we develop a hybrid framework to generate abundant synthetic examples from very few collected examples, and then we integrate them as the hybrid data for training the reliable student network.
\subsection{Motivation of the Hybrid Learning}
Formally, we denote the distribution of the collected data $\mathcal{D}_\mathrm{c}$ and synthetic data $\mathcal{D}_\mathrm{s}$ as $P$ and $Q$, respectively, while the distribution of the hybrid data $\mathcal{D}=\mathcal{D}_\mathrm{c}\cup\mathcal{D}_\mathrm{s}$ is represented as $U = \alpha P+(1-\alpha)Q$. Here $\alpha=\left|\mathcal{D}_\mathrm{c}\right| /\left(\left|\mathcal{D}_\mathrm{c}\right|+\left|\mathcal{D}_\mathrm{s}\right|\right)$ represents the proportion of collected examples in the hybrid data. In general, the synthetic and collected examples usually exhibit a significant distribution gap. This can cause substantial fluctuations during the training of the student network on hybrid data, ultimately leading to poor performance~\cite{wang2024generated}. Therefore, it is essential to align the distribution of synthetic data with that of collected data, thereby forming reliable hybrid data. Here, the synthetic data is generated under the supervision of the collected data, so we assume that synthetic data, collected data, and hybrid data have the same support set $\mathcal{X}$. Then, the distribution gap between the reliable hybrid data and synthetic data can be characterized by the Total Variation Distance (TVD), which is defined as
\begin{equation}
\mathrm{TVD}(U,Q)=\frac{1}{2}\sum_{\boldsymbol{x}\in\mathcal{X}}|U(\boldsymbol{x})-Q(\boldsymbol{x})|,
\end{equation}
where $U(\boldsymbol{x})\in(0,1)$ and $Q(\boldsymbol{x})\in(0,1)$ measure the distribution probability of $\boldsymbol{x}$ in the hybrid and synthetic data, respectively. Here $\mathrm{TVD}(\cdot,\cdot)\geq0$, and $\mathrm{TVD}(U,Q)=\frac{1}{2}\sum_{\boldsymbol{x}\in\mathcal{X}}|U(\boldsymbol{x})-Q(\boldsymbol{x})|\leq\frac{1}{2}\sum_{\boldsymbol{x}\in\mathcal{X}}(|U(\boldsymbol{x})|+|Q(\boldsymbol{x})|)=1$. Based on the triangle inequality~\cite{steerneman1983total} of TVD, we easily have that
\begin{equation}
\mathrm{TVD}(U, Q) \leq \mathrm{TVD}(U, P) + \mathrm{TVD}(P, Q).
\label{eq_4}
\end{equation}
Then, given that $U = \alpha P + (1 - \alpha)Q$ with parameter $\alpha$ controlling the weight of collected data, we can compute \( \mathrm{TVD}(U, P) \) as
\begin{equation}
\begin{aligned}
\text{TVD}(U, P) & =\frac{1}{2}\sum_{\boldsymbol{x}\in\mathcal{X}}|U(\boldsymbol{x})-P(\boldsymbol{x})| \\
& =\frac{1}{2}\sum_{\boldsymbol{x}\in\mathcal{X}}|\alpha P(\boldsymbol{x})+(1-\alpha)Q(\boldsymbol{x})-P(\boldsymbol{x})| \\
& =\frac{1}{2}(1-\alpha)\sum_{\boldsymbol{x}\in\mathcal{X}}|Q(\boldsymbol{x})-P(\boldsymbol{x})| \\
& =(1-\alpha) \text{TVD}\left(Q, P\right).
\end{aligned}
\end{equation}
By invoking the symmetry of TVD and Eq.~\eqref{eq_4}, we obtain
\begin{equation}
\mathrm{TVD}(U,Q)\leq(2-\alpha)\mathrm{TVD}(P,Q).
\label{theorem_gap}
\end{equation}
Here Eq.~\eqref{theorem_gap} reveals that the \textbf{high-quality synthetic data} $\mathcal{D}_\mathrm{s}$ and the \textbf{mix proportion} $\alpha$ are two critical factors influencing the distribution gap $\mathrm{TVD}(U,P)$.

The above observation inspires us to employ two modules to align the distribution of synthetic data with that of collected data, as shown in Fig.~\ref{fig_gandc}(c). In the teacher-guided generation module, we employ the teacher network to guide the GAN to enhance the quality of synthetic data, which solves its intrinsic issues when trained on the small and imbalanced collected data, including the overfitting of discriminator and imbalanced learning of generator:\\ 
\textbf{Discriminator Overfitting.} When trained with very few collected data, the discriminator is prone to be overconfident in determining fake examples, \textit{i.e.}, $\mathbb{E}_{\boldsymbol{x} \sim Q_{X}}[\mathcal{N}_\mathrm{D}(\boldsymbol{x})]$ tends to be 0. As a result, the gradient of $\mathcal{L}_{\mathrm{g}}$ in Eq.~\eqref{eq_gan}, which specialized in promoting generator to produce high-quality examples, may become ineffective, namely
\begin{equation}
\nabla_{\mathcal{N}_\mathrm{G}}\!\mathbb{E}_{\boldsymbol{x} \sim Q_{X}\!\!}\left[\log (1\!\!-\!\!\mathcal{N}_\mathrm{D}(\boldsymbol{x}))\right]\!\!=\!\!\mathbb{E}_{\boldsymbol{x} \sim Q_{X}}\!\! \left[-\frac{\nabla_{\mathcal{N}_\mathrm{G}}\!\mathcal{N}_\mathrm{D}(\boldsymbol{x})}{1 \!-\! \mathcal{N}_\mathrm{D}(\boldsymbol{x})}\right] \!\!\!\approx\!\!0,
\label{eq_0}
\end{equation}
as the parameters of $\mathcal{N}_\mathrm{D}$ and $\mathcal{N}_\mathrm{G}$ are independent of each other, and Eq.~\eqref{eq_0} is proved by~\cite{arjovsky2022towards}. Meanwhile, the discriminator also has a classifier that provides valuable category dependencies for the generator by precisely predicting input examples, and thus promoting the generator to generate classifiable examples. However, multi-class classification is more challenging than binary determination of true and fake. Given very few collected examples, the discriminator is difficult to learn powerful representations for its classifier to achieve precise classification.\\
\textbf{Imbalanced Generator Learning.} Given the optimal classifier $\Psi^{*}_{\mathrm{D}}$\footnote{$\Psi_{\mathrm{D}}^*\!\left(\!y^{+}\!\! \mid \boldsymbol{x}\right)\!\!\!=\!\!\!\frac{p(\boldsymbol{x}, y)}{p(\boldsymbol{x})+q(\boldsymbol{x})}, \Psi_{\mathrm{D}}^*\!\left(\!y^{-}\!\! \mid \boldsymbol{x}\right)\!\!\!=\!\!\!\frac{q(\boldsymbol{x}, y)}{p(\boldsymbol{x})+q(\boldsymbol{x})}$ (see Appendix).} of the discriminator, optimizing the generator to produce the classifiable examples\footnote{$\max_{\mathcal{N}_\mathrm{G}}[\mathbb{E}_{\boldsymbol{x}, y \!\sim Q_{X,\!Y}}[\log \Psi^{*}_{\mathrm{D}}(y^{+}\!\!\!\mid\!\!\boldsymbol{x})]\!-\!\!\mathbb{E}_{\boldsymbol{x}, y \!\sim Q_{X,\!Y}}[\log \Psi^{*}_{\mathrm{D}}(y^{-}\!\!\mid\!\!\boldsymbol{x})]]$.} is equivalent to
\begin{equation}
\max_{\mathcal{N}_\mathrm{G}} [\mathbb{E}_{\boldsymbol{x},y\sim Q_{X,Y}}\!\!\log (\frac{p(\boldsymbol{x}, y)}{q(\boldsymbol{x}, y)})]\!\!\Rightarrow\!\!\min_{\mathcal{N}_\mathrm{G}} \mathrm{KL}\!\left(Q_{X, Y} \| P_{X, Y}\!\right),
\label{eq_gl}
\end{equation}
where $\mathrm{KL}$ represents the Kullback-Leibler divergence, and the proof of Eq.~\eqref{eq_gl} is provided in Appendix. The above Eq.~\eqref{eq_gl} indicates that optimizing the generator will force the joint distribution $Q_{X, Y}$ of synthetic data toward the $P_{X, Y}$ of the imbalanced collected data, inevitably resulting in synthetic examples with poor diversity.

In student distillation, we properly inflate the collected examples to construct the hybrid data with a moderate mix proportion $\alpha$ for effectively training the student network. 
\subsection{Teacher-Guided Generation}
\label{sec_tgg}
In this section, we promote GAN to generate high-quality examples by solving its critical issues guided by the teacher network. To mitigate the discriminator overfitting, we design a feature integration mechanism to force the aggregation between the features of both real collected examples and fake synthetic examples. Specifically, we blend the boundaries between real and fake examples to increase the difficulty for the discriminator to accurately discriminate them, and thus preventing the discriminator from overconfidence, \textit{i.e.},
\begin{equation}
\begin{aligned}
\mathcal{L}_{\mathrm{blend}}\!\!=&\mathbb{E}_{\!\boldsymbol{x},y\sim \!P_{X,Y}\!,\boldsymbol{\hat{x}},y\sim \!Q_{X,Y}\!\!}[\mathbb{I}(p\!\!>\!\!q)(\!\left\|\Phi_{\mathrm{T}}(\boldsymbol{x})\!\!-\!\!\Phi_{\mathrm{D}}(\boldsymbol{\hat{x}})\right\|_{2}\!\!\\&+\left\|\Phi_{\mathrm{T}}(\boldsymbol{\hat{x}})\!\!-\!\!\Phi_{\mathrm{D}}(\boldsymbol{x})\right\|_{2})],
\end{aligned}
\label{eq_blend}
\end{equation}
where $\mathbb{I}(p>q)$ is an indicator function to control $\mathcal{L}_{\mathrm{blend}}$ be applied with a probability of $q$ and its value is 1 if $p>q$ and 0 otherwise ($p$ is sampled from $[0,1]$, $q$=0.7 and it is analyzed in Appendix). Meanwhile, we transfer the expressive features of the teacher network to enhance the representation ability of the discriminator, \textit{i.e.},
\begin{equation}
\begin{aligned}
\mathcal{L}_{\mathrm{trans}}=&\mathbb{E}_{\boldsymbol{x},y\sim P_{X,Y},\boldsymbol{\hat{x}},y\sim Q_{X,Y} }[(\left\|\Phi_{\mathrm{T}}(\boldsymbol{x})-\Phi_{\mathrm{D}}(\boldsymbol{x})\right\|_{2}\\&+\left\|\Phi_{\mathrm{T}}(\boldsymbol{\hat{x}})-\Phi_{\mathrm{D}}(\boldsymbol{\hat{x}})\right\|_{2})].
\end{aligned}
\label{eq_trans}
\end{equation}

To alleviate the imbalanced learning of the generator, we regularize the GAN training across all categories. During generator training, we dynamically update the class frequencies $\{n^{t}_{c}\}_{c=1}^{C}$ ($C$ represents the number of categories) at the beginning of iteration $t$ via the following exponential moving average function with a weight $\gamma \in[0,1]$, namely
\begin{equation}
n_{c}^{t} = (1-\gamma) n_{c}^{t-1} + \gamma\bar{n}_{c}^{t-1},
\label{eq_nc}
\end{equation}
where $\bar{n}_{c}^{t-1}$ is the number of synthetic examples belonging to class $c$ in iteration $t-1$, $n^{t}_{c}$ is initially set as a constant, and $\gamma$=0.5 (analyzed in Appendix). Then, each class frequency $n^{t}_{c}\in\{n^{t}_{c}\}_{c=1}^{C}$ is normalized as
\begin{equation}
\hat{n}_{c}^{t}=\frac{n^{t}_{c}}{\sum^{C}_{j=1} n^{t}_{j}}.
\end{equation}
Thereafter, the generator is regulated to produce balanced examples by minimizing the loss function:
\begin{equation}
\mathcal{L}_{\mathrm{reg}}=\sum_{c=1}^{C} \frac{\boldsymbol{p}_{\mathrm{T}}^{c} \log \left(\boldsymbol{p}_{\mathrm{T}}^{c}\right)}{\hat{n}_{c}^{t}},
\label{eq_reg}
\end{equation}
where $\boldsymbol{p}_{\mathrm{T}}\!\!\!=\!\!\!\mathbb{E}_{\boldsymbol{x}, y \sim Q_{X, Y}}[\mathrm{SoftMax}(\mathcal{N}_{\mathrm{T}}(\boldsymbol{x}))]$ is the average softmax vector output by the teacher network. The teacher network is well-trained on the original data, so it can precisely predict synthetic examples. In such a case, $\boldsymbol{p}_{T}^{c}$ can be regarded as the proportion of examples in category $c$ within the synthetic data. In Eq.~\eqref{eq_reg}, the generation of examples in a category $c$ with the lower (or higher) $\boldsymbol{p}_{\mathrm{T}}^{c}$ is adjusted by the larger (or smaller) $1/\hat{n}^{t}_{c}$.

The loss functions of discriminator and generator in our teacher-guided GAN are summarized as
\begin{equation}
\left\{\begin{array}{l}
\mathcal{L}_{\mathrm{D}}=\mathcal{L}_{\mathrm{adc\_d}}+\lambda_{\mathrm{d}}(\mathcal{L}_{\mathrm{blend}}+\mathcal{L}_{\mathrm{trans}}),\\
\mathcal{L}_{\mathrm{G}}=\mathcal{L}_{\mathrm{adc\_g}}+\lambda_{\mathrm{g}}\mathcal{L}_{\mathrm{reg}},
\end{array}\right.
\label{eq_total}
\end{equation}
where $\mathcal{L}_{\mathrm{adc\_d}}$ and $\mathcal{L}_{\mathrm{adc\_g}}$ are defined in Eq.~\eqref{eq_gan}, and the trade-off parameters $\lambda_{\mathrm{d}}>0$ and $\lambda_{\mathrm{g}}>0$.
\subsection{Student Distillation}
\label{sec_app_sl}
In the teacher-guided generation module, we successfully trained an effective GAN for generating high-quality synthetic examples, which are then combined with collected examples to construct the hybrid data $\mathcal{D}$ for training the student network. However, directly composing the limited collected examples with numerous synthetic examples will result in a small mix ratio $\alpha$ (\textit{i.e.}, a large distribution gap $\mathrm{TVD}(U, Q)$) to disturb the training of the student network. Therefore, we inflate the collected data via example repeating to enlarge the $\alpha$ from $\left|\mathcal{D}_\mathrm{c}\right| /\left(\left|\mathcal{D}_\mathrm{c}\right|+\left|\mathcal{D}_\mathrm{s}\right|\right)$ to $N\times\left|\mathcal{D}_\mathrm{c}\right| /\left(N\times\left|\mathcal{D}_\mathrm{c}\right|+\left|\mathcal{D}_\mathrm{s}\right|\right)$, where $N$ is the inflation factor. We adopt a moderate inflation factor of $N=\lfloor|\mathcal{D}_\mathrm{s}|/|\mathcal{D}_\mathrm{c}|\rfloor$ and further details are available in Extended Experiments.

Recent works~\cite{chen2021learning,tang2023distribution} indicate that the collected data usually contains many noisy examples, which may mislead the GAN to produce undesired synthetic examples with wrong labels. As a result, these potentially noisy examples will harm the training of the student network, particularly affecting its classifier. In DFKD, the teacher network is well-trained on the original data and possesses an accurate classifier. Recent studies~\cite{tang2023distribution,chen2022knowledge} show that the teacher's classifier contains useful category information regarding the original data. Therefore, we share the classifier of the teacher network with the student network. Then, we closely align the feature of the student network with that of the teacher network as follows:
\begin{equation}
\mathcal{L}_{\mathrm{align}} = \mathbb{E}_{\boldsymbol{x} \sim \mathcal{D}} \left[ \|\Phi_{\mathrm{S}}(\boldsymbol{x}) - \Phi_{\mathrm{T}}(\boldsymbol{x})\|_2 \right].
\end{equation}
By minimizing the $\mathcal{L}_{\mathrm{align}}$, the feature of the student network is closely aligned with that of the teacher network, and the aligned feature is inputted into the shared classifier can produce predictions as accurately as the teacher network. The student network did not use any example labels during the training process, thereby avoiding the negative impact of potentially noisy labels. The whole algorithm of our proposed HiDFD is given in Appendix.
\begin{table*}[t]
	\begin{center}
		\begin{adjustbox}{width=\linewidth}
			\begin{tabular}{cccc|cccccc|cccccccc}
				\hline
				\multirow{3}{*}{Dataset}      & \multirow{3}{*}{Arch}       & \multirow{3}{*}{ACC$^{\mathrm{T}}$} & \multirow{3}{*}{ACC$^{\mathrm{S}}$} & \multicolumn{6}{c|}{Generation-Based}                              & \multicolumn{8}{c}{Collection-Based}                                                                                                                                                           \\
				&                             &                                     &                                     & DAFL  & DDAD  & DI    & PRE   & CMI   & SSNet                      & \multicolumn{2}{c}{DeGAN}                                       & \multicolumn{2}{c}{DFND} & \multicolumn{2}{c}{KD$^{3}$}            & \multicolumn{2}{c}{HiDFD (ours)}                        \\
				&                             &                                     &                                     &       &       &       &       &       &                            & \multicolumn{1}{c}{$\rho$=0.1} & \multicolumn{1}{c}{$\rho$=1.0} & $\rho$=0.1  & $\rho$=1.0 & $\rho$=0.1 & $\rho$=1.0                 & $\rho$=0.1                 & $\rho$=1.0                 \\ \hline
				\multirow{3}{*}{CIFAR10}      & $\diamondsuit$ & 95.70                               & 95.20                               & 92.22 & 93.08 & 93.26 & 93.25 & 94.84 & \textbf{95.39}             & 90.39                          & 91.95                          & 48.82       & 85.82      & 65.70      & 93.37                      & 94.74                      & $\underline{\text{95.11}}$ \\
				& $\square$      & 94.07                               & 92.69                               & 86.92 & 90.85 & 85.27 & 91.82 & 88.49 & 92.00                      & 87.52                          & 90.37                          & 48.65       & 89.22      & 48.93      & 91.49                      & $\underline{\text{92.28}}$ & \textbf{93.14}             \\
				& $\triangle$    & 95.70                               & 92.69                               & 83.36 & 89.76 & 90.24 & 91.53 & 86.63 & 92.03                      & 86.40                          & 89.69                          & 49.48       & 90.60      & 65.10      & $\underline{\text{93.05}}$ & 92.90                      & \textbf{93.76}             \\ \hline
				\multirow{3}{*}{CIFAR100}     & $\diamondsuit$ & 78.05                               & 77.10                               & 74.47 & 73.64 & 61.32 & 74.19 & 77.04 & $\underline{\text{77.41}}$ & 53.20                          & 62.94                          & 21.45       & 64.73      & 26.96      & 72.90                      & 76.93                      & \textbf{78.35}             \\
				& $\square$      & 74.53                               & 72.28                               & 65.36 & 68.33 & 60.00 & 70.34 & 59.70 & 71.16                      & 53.97                          & 61.80                          & 23.48       & 63.90      & 21.27      & $\underline{\text{71.44}}$ & 71.26                      & \textbf{74.18}             \\
				& $\triangle$    & 78.05                               & 72.28                               & 45.28 & 68.59 & 61.07 & 67.49 & 61.80 & 72.38                      & 46.82                          & 56.44                          & 23.86       & 64.54      & 25.25      & 72.46                      & $\underline{\text{73.44}}$ & \textbf{75.65}             \\ \hline
				\multirow{3}{*}{CINIC}        & $\diamondsuit$ & 86.62                               & 85.09                               & 60.54 & 80.10 & 78.57 & 77.56 & 78.47 & 83.47                      & 57.59                          & 76.78                          & 24.53       & 80.94      & 39.35      & 82.68                      & $\underline{\text{85.62}}$ & \textbf{86.68}             \\
				& $\square$                      & 84.22                               & 83.28                               & 59.08 & 77.90 & 68.90 & 65.38 & 74.99 & 79.63                      & 54.36                          & 76.11                          & 29.53       & 77.41      & 29.88      & 78.18                      & $\underline{\text{81.92}}$ & \textbf{82.27}             \\
				& $\triangle$    & 86.62                               & 83.28                               & 44.62 & 77.63 & 59.52 & 63.23 & 75.46 & 80.30                      & 54.43                          & 74.40                          & 33.40       & 79.33      & 71.57      & 80.28                      & $\underline{\text{81.90}}$ & \textbf{82.88}             \\ \hline
				\multirow{3}{*}{\begin{tabular}[c]{@{}c@{}}Tiny-\\ImageNet\end{tabular}} & $\diamondsuit$ & 66.44                               & 64.87                               & 52.20 & 59.84 & 6.98  & 50.15 & 64.01 & 64.04                      &25.74                                & 49.11                               & 26.36       & 60.09      & 20.26      & 63.63                      & $\underline{\text{65.96}}$ & \textbf{66.61}             \\
				& $\square$      & 62.34                               & 61.55                               & 53.89 & 42.25 & 1.22  & 45.92 & 17.73 & 57.82                      &  23.13                              & 44.65                               & 25.39       & 58.47      & 24.26      & $\underline{\text{61.06}}$ & 60.46                      & \textbf{62.69}             \\
				& $\triangle$    & 66.44                               & 61.55                               & 52.46 & 44.20 & 2.27  & 47.22 & 20.57 & 59.16                      &21.09                                &48.12                                & 25.53       & 58.18      & 27.37      & $\underline{\text{61.98}}$ & 61.67                      & \textbf{65.27}             \\ \hline
				HAM                           & $\diamondsuit$ & 81.18                               & 79.64                               & 32.05 & 44.68 & 62.79 & 63.20 & 67.34 & 74.52                      & 34.75                               &  64.43                              & 27.55       & 62.59      & 64.10      & 68.44                      & $\underline{\text{77.08}}$ & \textbf{81.52}             \\ \hline
				ImageNet                      & $\diamondsuit$ & 73.27                               & 67.00                               & 1.92  & 1.46  & 1.14  & 1.60  & 1.84  & 5.74                       & 22.28                          & 43.96                          & 28.99       & 45.66      & 35.02      & 55.05                      & $\underline{\text{65.36}}$ & \textbf{66.89}             \\ \hline
			\end{tabular}
		\end{adjustbox}
	\end{center}
	\caption{Accuracies (in \%) of student networks trained by various methods on six image classification datasets. The columns ``$\text{ACC}^{\mathrm{T}}$'' and ``$\text{ACC}^{\mathrm{S}}$'' report the accuracies yielded by the teacher network and student network trained on the full original data, respectively. The best and the second-best results are highlighted in \textbf{bold} and \underline{underlined}, respectively. The notations $\diamondsuit$, $\square$, and $\triangle$ represent the teacher-student pairs ResNet34-ResNet18, ResNet34-VGG13, and VGG16-VGG13, respectively.}
	\label{table_class}
	\vspace{-10pt}
\end{table*}
\section{Experiments}
\label{sec_exp}
In this section, we employ various DNNs commonly utilized in DFKD methods~\cite{chen2021learning,tang2023distribution} and conduct intensive experiments on different benchmark datasets to evaluate the effectiveness of our HiDFD.
\subsection{Datasets and Implementation Details}
\label{sec_exp_dandi}
\textbf{Original Datasets.} We evaluate the effectiveness of our HiDFD on popular datasets, including CIFAR~\cite{krizhevsky2009learning}, CINIC~\cite{darlow2018cinic}, and TinyImageNet~\cite{le2015tiny}, which are widely used by existing DFKD methods~\cite{chen2019data,chen2021learning}. Additionally, we also conduct experiments on the large-scale ImageNet~\cite{deng2009imagenet} and the practical medical image dataset HAM~\cite{tsch2018ham}, which are challenging for existing DFKD methods.\\
\textbf{Collected Datasets.} When using CIFAR and CINIC as the original datasets, we search for examples from ImageNet. With TinyImageNet and ImageNet as the original datasets, we utilize WebVision~\cite{li2017webvision} as our source of collected data. Moreover, we collect examples from ISIC~\cite{codella2018skin} when using HAM as the original dataset. We follow~\cite{chen2021learning} and sample a part of examples from the corresponding dataset as collected data $\mathcal{D}_{\mathrm{c}}$. Here, we define the ratio between the collected data $\mathcal{D}_{\mathrm{c}}$ and original data $\mathcal{D}_{\mathrm{o}}$ as $\rho=|\mathcal{D}_{\mathrm{c}}|/|\mathcal{D}_{\mathrm{o}}|$. We construct small ($\rho$=$0.1$) and moderate ($\rho$=$1.0$) collected data for the experiments of collection-based DFKD methods. Notably, the original dataset is solely required for the pre-training of the teacher network. Detailed information regarding these datasets and the corresponding synthesized examples are provided in Appendix. 

\textbf{Implementation Details.} All student networks in our HiDFD employ SGD with weight decay as 5$\times$ 10$^{-4}$ and momentum as 0.9 as the optimizer. The student networks are trained over 240 epochs with a learning rate of 0.05, which is sequentially divided by 10 at the 150th, 180th, and 210th epochs. Meanwhile, the generator and discriminator in GAN utilize Adam for optimization with learning rates 1$\times$ 10$^{-4}$ and 4$\times$ 10$^{-4}$, respectively, and both of them are trained over 500 epochs. Additionally, the hyper-parameters in Eq.~\eqref{eq_total} are configured as $\lambda_{\mathrm{d}}=0.1$ and $\lambda_{\mathrm{g}}=0.1$.
\subsection{Experiments on Benchmark Datasets}
\label{sec_exp_cls}
In this section, we conduct comprehensive experiments on various benchmark datasets to evaluate the performance of our proposed HiDFD against SOTA generation-based~\cite{chen2019data,zhao2022dual,yin2020dreaming,binici2022robust,fang2021contrastive,yu2023data} and collection-based~\cite{addepalli2020degan,chen2021learning,tang2023distribution} DFKD methods. These methods are reproduced by using their official source codes.

Tab.~\ref{table_class} reports the results of the compared methods and our proposed HiDFD. Firstly, our proposed HiDFD using only a small quantity of collected examples ($\rho$=0.1) achieves comparable performance with those trained on the full original data. Secondly, when trained on the modestly sized collected data ($\rho$=1.0), our proposed HiDFD significantly outperforms compared methods on most tasks, especially on the challenging HAM and ImageNet. Thirdly, those generation-based methods, which utilize generative models to produce training examples without the supervision of real examples, tend to perform unsatisfactorily due to the deficiencies in their synthetic examples. These results demonstrate that our proposed HiDFD can train robust student networks by effectively generating training examples from limited real-world examples and properly utilizing all realistic examples.
\begin{table}[t]
	\vspace{-5pt}
	\centering
	\begin{adjustbox}{width=\linewidth}
		\begin{tabular}{clcc}
			\toprule
			Type                              & Algorithm & CIFAR10 & CIFAR100 \\ \hline
			\multirow{8}{*}{\begin{tabular}[c]{@{}l@{}}Teacher-\\Guided \\Generation\end{tabular}}    & w/o $\mathcal{L}_{\mathrm{blend}}$          & 92.87 (\textcolor{red}{$\downarrow$1.87})       & 74.18 (\textcolor{red}{$\downarrow$2.75})  \\ 
			& w/o $\mathcal{L}_{\mathrm{trans}}$    & 91.86 (\textcolor{red}{$\downarrow$2.88})         & 74.40 (\textcolor{red}{$\downarrow$2.53})  \\
			&  w/o $\mathcal{L}_{\mathrm{reg}}$     & 92.76 (\textcolor{red}{$\downarrow$1.98})         & 73.95 (\textcolor{red}{$\downarrow$2.98})  \\  
			& w/o $\mathcal{L}_{\mathrm{blend}}$, $\mathcal{L}_{\mathrm{trans}}$          & 91.10 (\textcolor{red}{$\downarrow$3.64})         & 71.02 (\textcolor{red}{$\downarrow$5.91})  \\
			& w/o $\mathcal{L}_{\mathrm{blend}}$, $\mathcal{L}_{\mathrm{reg}}$          & 90.77 (\textcolor{red}{$\downarrow$3.97})         & 71.83 (\textcolor{red}{$\downarrow$5.10})  \\
			& w/o $\mathcal{L}_{\mathrm{trans}}$, $\mathcal{L}_{\mathrm{reg}}$          & 91.42 (\textcolor{red}{$\downarrow$3.32})         & 72.10 (\textcolor{red}{$\downarrow$4.83})  \\
			& w/o $\mathcal{L}_{\mathrm{blend}}$, $\mathcal{L}_{\mathrm{trans}}$, $\mathcal{L}_{\mathrm{reg}}$          & 89.55 (\textcolor{red}{$\downarrow$5.19})         & 70.25 (\textcolor{red}{$\downarrow$6.68})  \\ \hline
			\multirow{8}{*}{\begin{tabular}[c]{@{}l@{}}Student \\Distillation\end{tabular}} 
			&  OFAKD~\cite{hao2024one}         & 92.88 (\textcolor{red}{$\downarrow$1.86})         & 70.86 (\textcolor{red}{$\downarrow$6.07})  \\			
			&  VKD~\cite{Hinton2015Distilling}        & 92.69 (\textcolor{red}{$\downarrow$2.05})         & 66.96 (\textcolor{red}{$\downarrow$9.97})  \\
			&  SemcKD~\cite{chen2021cross}      & 93.49 (\textcolor{red}{$\downarrow$1.25})         & 70.93 (\textcolor{red}{$\downarrow$6.00})  \\		
			&  CC~\cite{peng2019correlation}         & 92.63 (\textcolor{red}{$\downarrow$2.11})         & 69.52 (\textcolor{red}{$\downarrow$7.41})  \\		
			&  DKD~\cite{zhao2022decoupled}        & 92.95 (\textcolor{red}{$\downarrow$1.79})         & 68.25 (\textcolor{red}{$\downarrow$8.68})  \\   	
			&  RKD~\cite{park2019relational}      & 92.40 (\textcolor{red}{$\downarrow$2.34})         & 70.53 (\textcolor{red}{$\downarrow$6.40})\\   		
			& CATKD~\cite{guo2023class}         & 92.49 (\textcolor{red}{$\downarrow$2.25})   & 68.69 (\textcolor{red}{$\downarrow$8.24})  \\ 
			&  NKD~\cite{yang2023knowledge}       & 93.26 (\textcolor{red}{$\downarrow$1.48})         & 65.31 (\textcolor{red}{$\downarrow$11.62})\\ 		
			\hline
			\multicolumn{1}{c}{}  &HiDFD (ours)                                   &94.74         &76.93   \\ \hline
		\end{tabular}
	\end{adjustbox}
	\caption{Accuracies (in \%) of ablation studies.}
	\label{table_ab}
	\vspace{-10pt}
\end{table}
\begin{figure*}[t]
	\centering
	\includegraphics[scale=0.3]{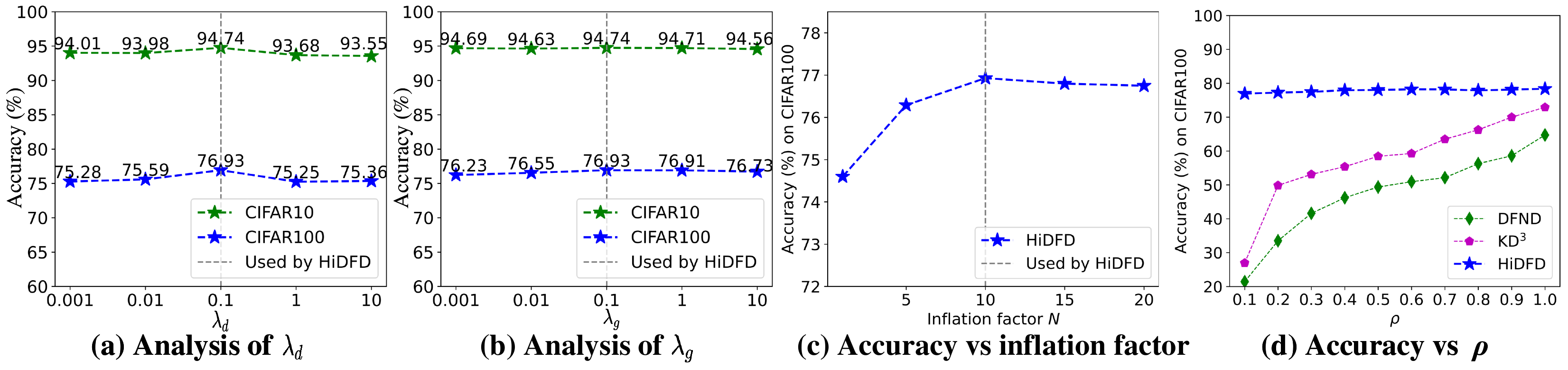}
	\caption{Parametric sensitivities of (a) $\lambda_{\mathrm{d}}$ and (b) $\lambda_{\mathrm{g}}$ in Eq.~\eqref{eq_total}. Accuracies (in \%) of the student networks trained with collected data with (c) varying inflation factors and (d) various quantities.}
	\label{fig_ps}
\end{figure*}
\subsection{Ablation Studies \& Parametric Sensitivities}
\label{sec_exp_ab}
In this section, we evaluate the effectiveness of our method with a small collected data ($\rho$=0.1), where CIFAR and ImageNet serve as the original and collected datasets, respectively. Moreover, ResNet34 and ResNet18 are used as the teacher network and student network, respectively. \\
\textbf{Ablation Studies.} We evaluate three key operations ($\mathcal{L}_{\mathrm{blend}}$, $\mathcal{L}_{\mathrm{trans}}$, and $\mathcal{L}_{\mathrm{reg}}$) in teacher-guided generation and the classifier-sharing-based strategy in the student distillation. The experimental results are reported in Tab.~\ref{table_ab}, and the contributions of these components are analyzed as follows:\\
\textbf{1) Teacher-Guided Generation.} The feature blending $\mathcal{L}_{\mathrm{blend}}$ in Eq.~\eqref{eq_blend} and feature transferring $\mathcal{L}_{\mathrm{trans}}$ in Eq.~\eqref{eq_trans} for preventing the overfitting of discriminator and enhancing its representation ability. Meanwhile, the generator regulation $\mathcal{L}_{\mathrm{reg}}$ in Eq.~\eqref{eq_reg} is also essential for maintaining the balanced training of the generator. Therefore, the omission of any components among them leads to a noticeable reduction in the performance of the student network. Particularly, training the student network only on synthetic examples without any guidance from the teacher network results in the poorest performance (as shown in the term ``w/o $\mathcal{L}_{\mathrm{blend}}$, $\mathcal{L}_{\mathrm{trans}}$, $\mathcal{L}_{\mathrm{reg}}$''). These results indicate the importance of these operations for robust GAN training with limited collected examples, thereby generating high-quality examples for training reliable student networks. \\
\textbf{2) Student Distillation.} We examine the impact of replacing the classifier-sharing-based feature alignment with traditional KD methods~\cite{Hinton2015Distilling,chen2021cross}. Both the student networks are trained on the hybrid data composed of collected and synthetic examples. We can find that the student networks trained by these methods generally achieve suboptimal performance due to their inability to effectively handle the potentially noisy examples among the hybrid data. These results highlight the suitability of our training strategy for reliable student networks in the data-free distillation scenarios.\\
\textbf{Parametric Sensitivity.} There are two tuning parameters in our HiDFD, including $\lambda_{\mathrm{d}}$ and $\lambda_{\mathrm{g}}$ in Eq.~\eqref{eq_total}. To analyze the sensitivities, we individually vary each parameter while keeping the others constant during training. The accuracies of the corresponding student networks are shown in Fig.~\ref{fig_ps}(a) and Fig.~\ref{fig_ps}(b). Despite the large fluctuations in these parameters, where $\lambda_{\mathrm{d}}$, $\lambda_{\mathrm{g}}\in$ \{0.001,0.01, 0.1, 1, 10\}, the accuracy curve of the student network remains relatively stable. These results indicate the robustness of our HiDFD against parameter variations. Additionally, the student network achieved the best performance when $\lambda_{\mathrm{d}}=\lambda_{\mathrm{g}}=0.1$, so we adopted such parameter configuration in our method.
\subsection{Extended Experiments}
\label{sec_exp_extend}
\textbf{Experiments with Various Backbones.} We evaluate our HiDFD across many widely used teacher-student pairs to assess its adaptability to different networks. The results are shown in Tab.~\ref{table_extent}, we can observe that our HiDFD consistently achieves satisfactory performance across different teacher-student pairs, where both the trained students perform comparably to those trained on the original data.\\
\textbf{Experiments with Varying Inflation Factors.} We report the accuracies of the student networks trained on collected data with various inflation factors in Fig.~\ref{fig_ps}(c). The student network performs better with increasing $N$, and the best accuracy is observed at $N$=10. Furthermore, excessive inflation may reduce the diversity brought by synthetic data, so that the student network encounters performance degradation when $N \textgreater$10. Therefore, we adopt a moderate inflation factor of $N=\lfloor|\mathcal{D}_\mathrm{s}|/|\mathcal{D}_\mathrm{c}|\rfloor$. These experiments demonstrate that appropriately inflating the collected examples, which are crucial for reducing the distribution gap between synthetic and collected data, can effectively improve the performance of the student network.\\
\textbf{Experiments on Collected Data with Various Data Quantities.} We explore the impact of varying the volume of collected data on the performance of student networks, with $\rho$ values ranging from 0.1 to 1. As shown in Fig.~\ref{fig_ps}(d), student networks trained by the compared collection-based DFKD methods~\cite{chen2021learning,tang2023distribution} tend to underperform with small values of $\rho$. Conversely, our HiDFD consistently achieves satisfactory performance across a spectrum of $\rho$ values. These results further demonstrate the effectiveness of our HiDFD in training reliable student networks leveraging limited collected data.
\begin{table}
	\begin{center}
		    \begin{adjustbox}{width=\linewidth}
	\begin{tabular}{cccccc}
	\hline
	Dataset                   & Teacher   & Student    & $\text{ACC}^{\mathrm{S}}$   & HiDFD &\textcolor{green}{$\uparrow$} or \textcolor{red}{$\downarrow$}    \\ \hline
	\multirow{4}{*}{CIFAR10}  & ResNet32$\times$4  & ResNet110    & 93.37 & 95.04 & \textcolor{green}{$\uparrow$1.67} \\ 
	& ResNet32$\times$4  & ShuffleNet & 93.23 & 93.62 & \textcolor{green}{$\uparrow$0.39} \\ 
	& ResNet110$\times$2 & ResNet116  & 93.21 & 94.83 & \textcolor{green}{$\uparrow$1.62} \\  
	& ResNet110$\times$2 & WRN40$\times$2      & 94.86 & 95.35 & \textcolor{green}{$\uparrow$0.49} \\ \hline
	\multirow{4}{*}{CIFAR100} & ResNet32$\times$4  & ResNet110    & 74.31 & 75.69 & \textcolor{green}{$\uparrow$1.38} \\
	& ResNet32$\times$4  & ShuffleNet & 72.60 & 75.03 & \textcolor{green}{$\uparrow$2.43} \\ 
	& ResNet110$\times$2 & ResNet116  & 74.46 & 74.49 & \textcolor{green}{$\uparrow$0.03} \\ 
	& ResNet110$\times$2 & WRN40$\times$2      & 76.31 & 75.65 & \textcolor{red}{$\downarrow$0.66} \\ \hline
\end{tabular}
	    \end{adjustbox}
	\end{center}
	\caption{Accuracies (in \%) of various networks trained by our method ($\rho$=1.0).}
	\label{table_extent}
	\vspace{-10pt}
\end{table}
\section{Conclusion}
In this paper, we proposed a new data-free distillation approach termed HiDFD to train the student networks on the hybrid data comprising high-quality synthetic examples and scarce collected examples, which well meets practical requirements. Our investigation reveals that bridging the distribution gap between the hybrid and synthetic data is crucial for training reliable student networks, and it implies that the quality of synthetic data and the weight of collected data are two key factors in reducing this gap. This observation inspired us to propose a novel hybrid distillation framework, where the teacher-guided generation module can effectively generate high-quality synthetic examples from the limited collected data by leveraging the teacher network to guide the GAN training process, and the student distillation module properly enhances the influence of collected examples within the hybrid data by inflating their frequency. Consequently, we can naturally define a classifier-sharing-based feature alignment to distill the student network, and we achieve state-of-the-art performance using significantly fewer examples than existing methods. The limitations and broader impacts of our HiDFD are discussed in Appendix.
\section{Acknowledgments}
This research is supported by NSF of China (Nos: 62336003, 12371510), and NSF for Distinguished Young Scholar of Jiangsu Province (No: BK20220080).
\bibliographystyle{aaai25}
\bibliography{aaai25}
\clearpage
\end{document}


\maketitle
\appendix
\renewcommand{\thefigure}{A-\arabic{figure}}
\renewcommand{\thetable}{A-\arabic{table}}
\renewcommand{\thealgorithm}{A-\arabic{algorithm}}
\setcounter{figure}{0}
\setcounter{table}{0}
\begin{table*}[t]
	\centering
	\caption{Details of the benchmark datasets used in our experiment, the items with the prefix ``\#" denotes the number of the corresponding item.}
	\setlength{\tabcolsep}{1.5mm}
	{
		\begin{tabular}{cccccc}
			\hline
			Dataset   &Type   & Image size  &\#classes & \#train & \#test \\ \hline
			CIFAR10   &Original data   & 32$\times$32             & 10        & 50,000  & 10,000 \\ 
			CIFAR100  &Original data   & 32$\times$32              & 100       & 50,000  & 10,000 \\ 
			CINIC     &Original data   & 32$\times$32             & 10        & 90,000  & 90,000 \\ 
			TinyImageNet&Original data  & 64$\times$64              & 200       & 100,000 & 10,000 \\ 
			HAM &Original data & 224$\times$224              & 7       & 8,000 & 2,000 \\ 
			ISIC &Collected data & 224$\times$224              & 8       & 20,000 & 5,000 \\ 
			ImageNet &Original\&collected data & 224$\times$224              & 1,000       & 1,281,167 & 50,000 \\
			WebVision &Collected data & 224$\times$224              & 1,000       & 980,449 & N/A \\ \hline
	\end{tabular}}
	\label{table_dataset}
\end{table*}
\section{Proofs}
\label{sup_proof}
\subsection{The Optimal Classifier of Discriminator}
\label{sup_proof1}
For a fixed generator, the optimal classifier $\Psi_{\mathrm{D}}^*$ of the discriminator in our employed Auxiliary Discriminative Classifier GAN (ADCGAN) can be formatted as follows:
\begin{equation}
\begin{aligned}
\Psi_{\mathrm{D}}^*\left(y^{+} \mid \boldsymbol{x}\right)=\frac{p(\boldsymbol{x}, y)}{p(\boldsymbol{x})+q(\boldsymbol{x})}, \\  \Psi_{\mathrm{D}}^*\left(y^{-} \mid \boldsymbol{x}\right)=\frac{q(\boldsymbol{x}, y)}{p(\boldsymbol{x})+q(\boldsymbol{x})}.
\end{aligned}
\end{equation}
\begin{proof}
	\begin{equation}
	\begin{aligned}
	\max_{\Psi_{\mathrm{D}}}\mathbb{E}_{\boldsymbol{x},y\sim P_{X,Y}}[\log \Psi_{\mathrm{D}}(y^+|\boldsymbol{x})] +\\ \mathbb{E}_{\boldsymbol{x},y\sim Q_{X,Y}}[\log \Psi_{\mathrm{D}}(y^-|\boldsymbol{x})] \\
\Rightarrow\max_{\Psi_{\mathrm{D}}}\mathbb{E}_{\boldsymbol{x},y\sim P^m_{X,Y}}[\log \Psi_{\mathrm{D}}(y|\boldsymbol{x})],
	\end{aligned}
	\end{equation}
	with $p^m(\boldsymbol{x},y^+)=\frac{1}{2}p(\boldsymbol{x},y)$, $p^m(\boldsymbol{x},y^-)=\frac{1}{2}q(\boldsymbol{x},y)$, and $p^m(\boldsymbol{x})=\sum_y p^m(\boldsymbol{x},y)=\frac{1}{2}p(\boldsymbol{x})+\frac{1}{2}q(\boldsymbol{x})$.
	\begin{IEEEeqnarray}{rCl}
	&\Rightarrow&\max_{\Psi_{\mathrm{D}}}\mathbb{E}_{\boldsymbol{x}\sim P^m_{X}}\mathbb{E}_{y\sim P^m_{Y|X}}[\log \Psi_{\mathrm{D}}(y|\boldsymbol{x})]\\&\Rightarrow&\min_{\Psi_{\mathrm{D}}}\mathbb{E}_{\boldsymbol{x}\sim P^m_{X}}\mathbb{E}_{y\sim P^m_{Y|X}}[-\log \Psi_{\mathrm{D}}(y|\boldsymbol{x})]\\
	&\Rightarrow&\min_{\Psi_{\mathrm{D}}}\!\!\mathbb{E}_{\boldsymbol{x}\sim\! P^m_X}[H(p^m(y|\boldsymbol{x}))\!\!+\!\!\mathrm{KL}(p^m\!(y|\boldsymbol{x})\|\Psi_{\mathrm{D}}(y|\boldsymbol{x}))] \\
	&\Rightarrow& \Psi_{\mathrm{D}}^*(y|\boldsymbol{x})=\arg\min_{\Psi_{\mathrm{D}}}\mathrm{KL}(p^m(y|\boldsymbol{x})\|\Psi_{\mathrm{D}}(y|\boldsymbol{x}))\\&=&p^m(y|\boldsymbol{x})=\frac{p^m(\boldsymbol{x},y)}{p^m(\boldsymbol{x})}
	\end{IEEEeqnarray}
	Therefore, the optimal discriminative classifier of ADC-GAN has the form of $\Psi_{\mathrm{D}}^*(y^+|\boldsymbol{x})=\frac{p^m(\boldsymbol{x},y^+)}{p^m(\boldsymbol{x})}=\frac{p(\boldsymbol{x},y)}{p(\boldsymbol{x})+q(\boldsymbol{x})}$ and $\Psi_{\mathrm{D}}^*(y^-|\boldsymbol{x})=\frac{p^m(\boldsymbol{x},y^-)}{p^m(\boldsymbol{x})}=\frac{q(\boldsymbol{x},y)}{p(\boldsymbol{x})+q(\boldsymbol{x})}$ that conclude the proof.
\end{proof}
\subsection{Proof of Eq. (7)}
\label{sup_proof2}
Given the optimal classifier of the discriminator, at the equilibrium point, encouraging the generator to produce easily classifiable examples of our employed ADCGAN is equivalent to
\begin{equation}
\begin{aligned}
&\max_{\mathcal{N}_\mathrm{G}}[\mathbb{E}_{\boldsymbol{x}, y \sim Q_{X, Y}}[\log \Psi^{*}_{\mathrm{D}}(y^{+}\mid\boldsymbol{x})]\\&-\mathbb{E}_{\boldsymbol{x}, y \sim Q_{X, Y}}[\log \Psi^{*}_{\mathrm{D}}(y^{-}\mid\boldsymbol{x})]]\\&\Rightarrow\min _{\mathcal{N}_\mathrm{G}} \mathrm{KL}\left(Q_{X, Y} \| P_{X, Y}\right),
\end{aligned}
\end{equation}
\begin{proof}
	\begin{IEEEeqnarray}{rCl}
	&&\max_{\mathcal{N}_\mathrm{G}} \mathbb{E}_{\boldsymbol{x},y\sim Q_{X,Y}}\left[\log \Psi_{\mathrm{D}}^*(y^+|\boldsymbol{x})\right]\\&& - \mathbb{E}_{\boldsymbol{x},y\sim Q_{X,Y}}\left[\log \Psi_{\mathrm{D}}^*(y^-|\boldsymbol{x})\right]\\
	&\Rightarrow&\max_{\mathcal{N}_\mathrm{G}} \mathbb{E}_{\boldsymbol{x},y\sim Q_{X,Y}}\left[\log \frac{p(\boldsymbol{x},y)}{p(\boldsymbol{x})+q(\boldsymbol{x})}\right] \\&&- \mathbb{E}_{\boldsymbol{x},y\sim Q_{X,Y}}\left[\log \frac{q(\boldsymbol{x},y)}{p(\boldsymbol{x})+q(\boldsymbol{x})}\right] \\
	&\Rightarrow&\min_{\mathcal{N}_\mathrm{G}}\mathbb{E}_{\boldsymbol{x},y\sim Q_{X,Y}}\left[\log \frac{q(\boldsymbol{x},y)}{p(\boldsymbol{x},y)}\right]\\&\Rightarrow& \min_{\mathcal{N}_\mathrm{G}} \mathrm{KL}(Q_{X,Y}\|P_{X,Y})
	\end{IEEEeqnarray}
\end{proof}
\begin{figure*}[t]
	\centering
	\includegraphics[scale=0.43]{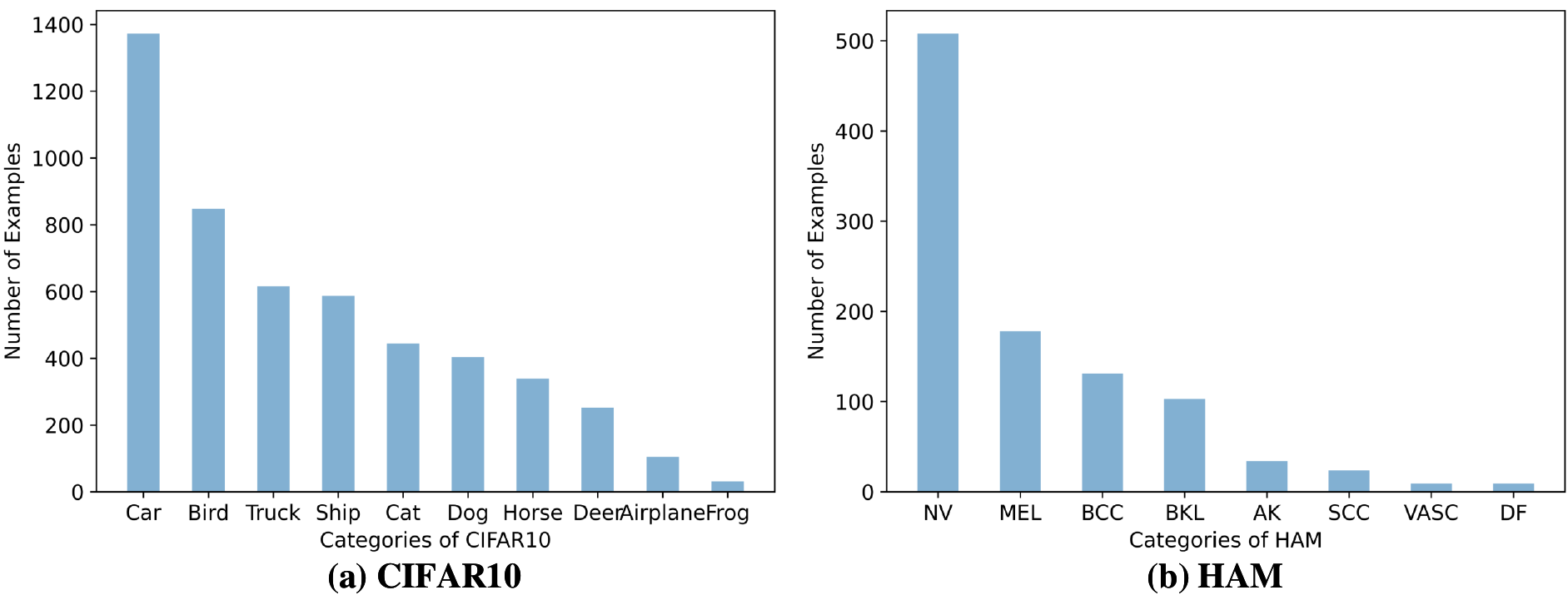}
	\caption{The number of examples per category in the collected data ($\rho=0.1$), where the original data is (a) CIFAR10 and (b) HAM, respectively.}
	\label{fig_isic}
\end{figure*}
\begin{algorithm}[t]
	\caption{Hybrid Data-Free Distillation}
	\begin{algorithmic}[1]
		\REQUIRE Pre-trained teacher network $\mathcal{N}_{\mathrm{T}}$, limted collected data $\mathcal{D}_{\mathrm{c}}$.
		\STATE Initialize the discriminator $\mathcal{N}_{\mathrm{D}}$ and generator $\mathcal{N}_{\mathrm{G}}$ in GAN, and the small student network $\mathcal{N}_{\mathrm{S}}$;
		\STATE \textbf{Module 1: Teacher-Guided Generation}
		\REPEAT 
		\STATE Sample the noise-label pair $(\boldsymbol{z}, y)$ to generate synthetic example $\mathcal{N}_\mathrm{G}(\boldsymbol{z}, y)$;
		\STATE Mitigate the overfitting of discriminator $\mathcal{N}_{\mathrm{D}}$ via blending $\mathcal{L}_{\mathrm{blend}}$ and transferring $\mathcal{L}_{\mathrm{trans}}$ operations in feature integration;
		\STATE Calculate the class frequency $\{\hat{n}_{c}\}_{c=1}^{C}$ of synthetic examples;
		\STATE Regularize the training of generator $\mathcal{N}_{\mathrm{G}}$ across all categories;
		\STATE Optimize the GAN via Adam;
		\UNTIL {convergence}
		\STATE \textbf{Module 1: Student Distillation}
		\STATE Generate abundant high-quality synthetic examples and inflate collected examples to construct the hybrid data $\mathcal{D}$; 
		\REPEAT 
		\STATE Sample the example $\boldsymbol{x} \in \mathcal{D}$ and input it into the teacher network $\mathcal{N}_{\mathrm{T}}$ and student network $\mathcal{N}_{\mathrm{S}}$ to obtain the features $\Phi_{\mathrm{T}}(\boldsymbol{x})$ and $\Phi_{\mathrm{S}}(\boldsymbol{x})$, respectively.
		\STATE Align the features $\Phi_{\mathrm{T}}(\boldsymbol{x})$ and $\Phi_{\mathrm{S}}(\boldsymbol{x})$ via $\mathcal{L}_{\mathrm{align}}$;
		\STATE Optimize the student network $\mathcal{N}_{\mathrm{S}}$ via SGD;
		\UNTIL {convergence}
		\ENSURE Lightweight student network $\mathcal{N}_{\mathrm{S}}$.
	\end{algorithmic}
	\label{alg_1}
\end{algorithm}
\section{Detailed Information of Benchmark Datasets}
\label{sup_dataset}
\subsection{Comprehensive Dataset Overview}
In Table~\ref{table_dataset}, we introduce the critical information of the benchmark datasets used in our experiment, including the image size, the number of categories, and the number of images in training and test sets. First, we can observe that the large-scale ImageNet contains a large number of examples, when using it as the original data, it is difficult to collect sufficient examples (more than 1 million) from the real-world. Therefore, it is essential to explore an efficient data-free distillation method that requires only a small amount of collected data. Second, a series of benchmark datasets contain 224$\times$224 sized high-resolution images, the generation-based methods that without rely on real-world data are hard to generate high-quality examples to provide effective information for training the student network.

Moreover, Figure~\ref{fig_isic} shows the number of collected examples in each category when the original data is the natural image datasets CIFAR10 and practical medical image dataset HAM, respectively. We can observe that the collected examples exhibit imbalanced class distribution, with several categories accounting for the majority of examples and other categories containing only a few examples. Therefore, our proposed data-free distillation method is very practical.
\begin{figure*}[t]
	\centering
	\includegraphics[scale=0.3]{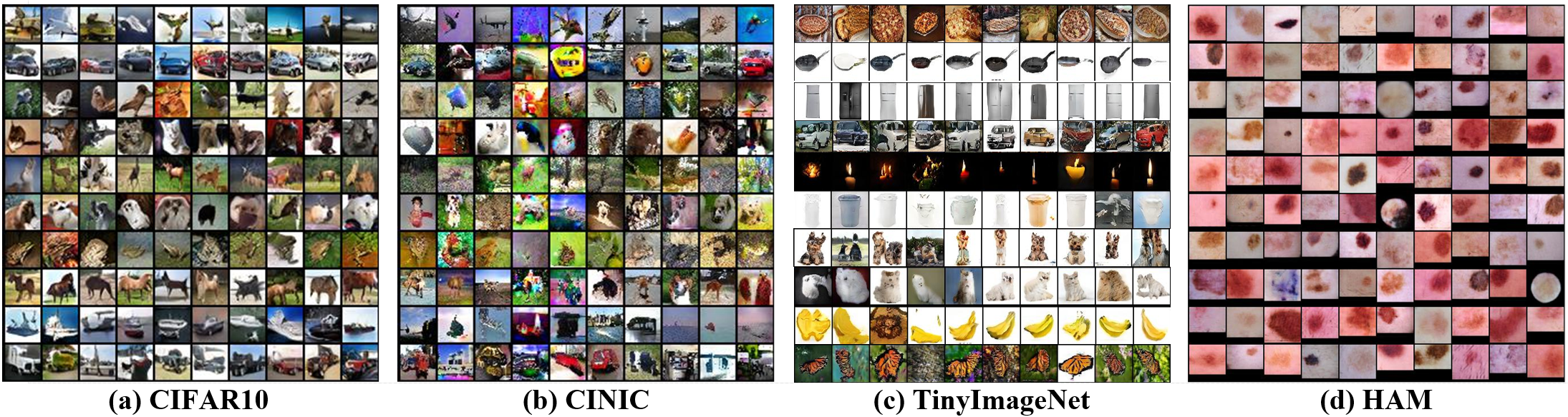}
	\caption{Visualization of synthetic examples for the original tasks, including (a) CIFAR10, (b) CINIC, (c) TinyImageNet, and (d) HAM.}
	\label{fig_vis_cifar10}
\end{figure*}
\subsection{Visualization of Synthetic Examples}
In Figure~\ref{fig_vis_cifar10}, we show synthetic examples produced by the GAN trained on limited collected data using our teacher-guided generation module. Specifically, we generate synthetic instances for four original datasets, including CIFAR10, CINIC, TinyImageNet, and HAM. Despite these original datasets having significantly different image sizes (ranging from 32$\times$32 to 224$\times$224), the corresponding synthetic examples consistently exhibit high quality. These visual results demonstrate the effectiveness of our approach in leveraging the teacher network to address critical issues of GAN training on limited collected data. By doing so, our method can train reliable student networks on abundant high-quality synthetic examples.
\section{Algorithm}
\label{sup_alg}
The detailed training algorithm of our proposed HiDFD is summarized in Algorithm~\ref{alg_1}. Our HiDFD contains two primary modules to train a reliable student network only using a small number of collected examples, \textit{i.e.}, the teacher-guided generation and student distillation. In the teacher-guided generation, the GAN is trained on the limited collected data under the guidance of the teacher network, where the critical issues, \textit{i.e.}, overfitting of the discriminator and imbalanced learning of the generator, are effectively resolved. In the student network, the collected examples are properly inflated via repeating and combined with sufficient high-quality synthetic examples to construct the hybrid data. Then, the reliable student network can naturally train on the hybrid data via the effective classifier-sharing-based feature alignment strategy.
\section{Limitations and Broader Impacts}
\label{sup_impact}
\subsection{Limitations}
The proposed method can train reliable student networks using very few collected examples. Compared with previous methods, we have reduced the data requirement by 99\%, making our method suitable for practical applications. In general, the effectiveness of the proposed method depends on the quality and representativeness of collected data to some extent. If this collected data does not sufficiently represent the broader dataset or contains biases, the generated synthetic examples and the trained student network may inherit these flaws. In practice, collecting fewer representative examples in real-world applications is relatively easy. Therefore, we believe that these limitations can be overcome well.
\begin{figure*}[t]
	\centering
	\includegraphics[scale=0.3]{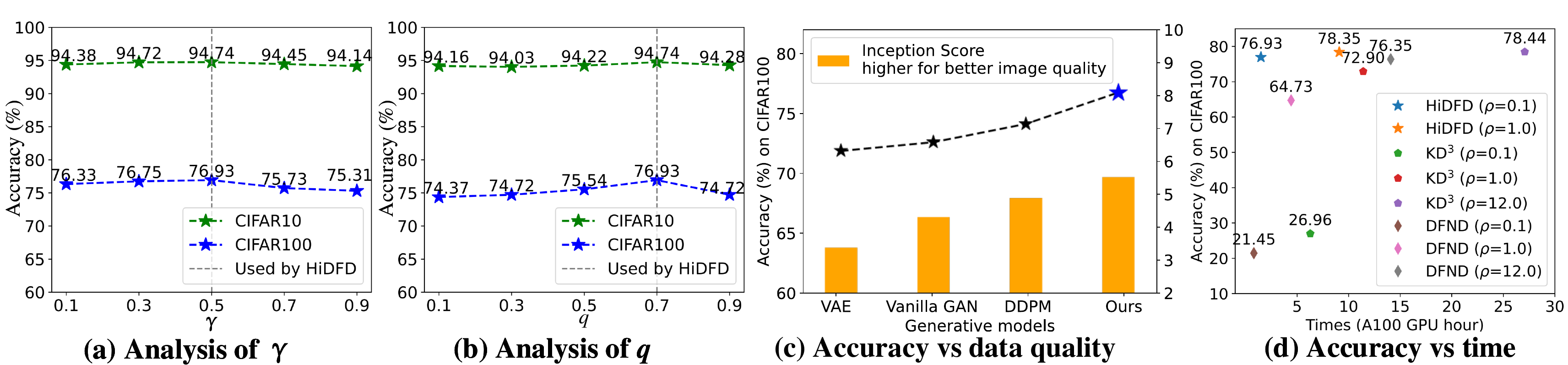}
	\caption{Parametric sensitivities of (a) $\gamma$ in Eq. (12) and (b) $q$ in Eq. (9). Accuracies (in \%) of the student networks trained with (c) synthetic examples generated by various generative models. (d) shows the accuracies and training times of various DFKD methods.}
	\label{fig_ps}
\end{figure*}
\subsection{Broader Impacts}
This paper proposed a novel data-free knowledge distillation method called HiDFD, which can train a compact and reliable student network using very few collected examples. In general, the proposed HiDFD could have the following positive impacts: 1) HiDFD eliminates the need for original training data required by traditional knowledge distillation methods, so it can help preserve data privacy for users; 2) HiDFD effectively compresses the large pre-trained models (\textit{i.e.}, the teacher networks) into smaller and faster models (\textit{i.e.}, the student networks) that are resource-efficient and suitable for deployment on devices with limited capabilities; 3) HiDFD focuses on the classification tasks, which underpin many practical downstream tasks like object detection and segmentation, suggesting its wide applicability; and 4) HiDFD is compatible with different DNNs (\textit{e.g.}, ResNet and VGG). 

Although HiDFD has few negative social impacts, when it compresses the large models of many Artificial Intelligence (AI) technologies and enables these compressed models in practical applications, the proposed HiDFD can be used for good and also for harm, depending on human intent. This actually falls into the general ethical debate on whether AI is good or not. 

In conclusion, we believe our proposed work can be beneficial to society since many important real-world applications need compact and reliable models that stand to benefit from HiDFD when the available real-world data is limited.
\section{Additional Experiments}
\subsection{Additional Parametric Sensitivities}
There are also two tuning parameters $q$ and $\gamma$ in Eq.~(9) and $\gamma$ in Eq.~(12), respectively. To analyze their sensitivities, we individually vary each parameter while keeping the others constant during training. The accuracies of the corresponding student networks are shown in Figure~\ref{fig_ps}. Despite the large fluctuations in these parameters, where $q$, $\gamma \in$ \{0.1, 0.3, 0.5, 0.7, 0.9\}, the accuracy curve of the student network remains relatively stable. These results indicate the robustness of our HiDFD against parameter variations. Additionally, the student network achieved the best performance when $q=0.7$ and $\gamma=0.5$, so we adopted such parameter configuration in our method.
\subsection{Experiments on Various Collected Data}
We compare the quality of synthetic examples produced by different generative models trained with the limited collected data and report the accuracies of the corresponding student networks in Figure~\ref{fig_ps}(c). Here, the higher-quality synthetic data consistently promotes a better student network, which demonstrates that improving the quality of synthetic examples can effectively improve the performance of the student network.

Moreover, we compared the training time of our method with other collection-based DFKD methods, as depicted in Figure~\ref{fig_ps}(d). We can observe that our method can train a student network with satisfactory performance within a few hours on an A100 GPU. These results further demonstrate the effectiveness of our HiDFD in training reliable student networks leveraging limited collected data.